\documentclass[letterpaper, 10pt, conference]{ieeeconf}  

\IEEEoverridecommandlockouts                              

\usepackage{cite}
\usepackage{graphicx}  
\usepackage[bookmarks=true, hidelinks]{hyperref}
\usepackage{rotating}  
\usepackage[normalem]{ulem}  
\usepackage{colortbl}
\usepackage{xcolor} 
\usepackage{amsmath}
\usepackage{amssymb}
\usepackage{amsbsy}
\usepackage{tabularx}
\usepackage{multirow}
\usepackage{hhline}
\usepackage{algorithm}
\usepackage[noend]{algpseudocode}

\usepackage{soul}

\newcommand{\zackory}[1]{\textcolor{purple}{{Zackory: #1}}}

\newcommand{\vc}[1]{\ensuremath{\mathbf{#1}}}

\newcommand{\etal}{{et~al.\ }}

\newcommand{\ie}{i.e.\ }

\long\def\ignorethis#1{}

\title{\LARGE \bf
Learning to Collaborate from Simulation for Robot-Assisted Dressing
}

\author{Alexander~Clegg$^{*,1,2}$,
        Zackory~Erickson$^{1}$,
        Patrick~Grady$^{1}$,
        Greg~Turk$^{1}$,
        Charles~C.~Kemp$^{1}$,
        and~C.~Karen~Liu$^{1,3}$
\thanks{\tt\small *alexanderwclegg@gmail.com}
\thanks{$^{1}$Georgia Institute of Technology}%
\thanks{$^{2}$Facebook Artificial Intelligence Research}%
\thanks{$^{3}$Stanford University}%
\thanks{This work was supported by NSF Graduate Research Fellowship under Grant No.DGE-1650044, NSF award IIS-1514258, and AWS Cloud Credits for Research.}%
}

\begin{document}

\maketitle
\thispagestyle{empty}
\pagestyle{empty}

\begin{abstract}



We investigated the application of haptic feedback control and deep reinforcement learning (DRL) to robot-assisted dressing. Our method uses DRL to simultaneously train human and robot control policies as separate neural networks using physics simulations. In addition, we modeled variations in human impairments relevant to dressing, including unilateral muscle weakness, involuntary arm motion, and limited range of motion. Our approach resulted in control policies that successfully collaborate in a variety of simulated dressing tasks involving a hospital gown and a T-shirt. In addition, our approach resulted in policies trained in simulation that enabled a real PR2 robot to dress the arm of a humanoid robot with a hospital gown. We found that training policies for specific impairments dramatically improved performance; that controller execution speed could be scaled after training to reduce the robot's speed without steep reductions in performance; that curriculum learning could be used to lower applied forces; and that multi-modal sensing, including a simulated capacitive sensor, improved performance. 
\end{abstract}


\section{Introduction}
It becomes ever more likely that robots will be found in homes and businesses, \emph{physically} interacting with the humans they encounter. With this in mind, researchers have begun preparing robots for the physical interaction tasks which they will face in a human world. Dressing tasks present a multitude of privacy, safety, and independence concerns which strongly motivate the application of robotic assistance \cite{adls_1990}. However, clothing exhibits complex dynamics and often occludes the body, making it difficult to accurately observe the task state and predict the results of planned interactions. These challenges are compounded by the risk of injuring people and the difficulty of recruiting participants, limiting the potential for robot exploration and learning. Learning from physics simulations presents a promising alternative that might enable robots to more freely explore and learn from vast amounts of data without putting people at risk.


\begin{figure}[t!]
\centering
\includegraphics[width=0.395\textwidth]{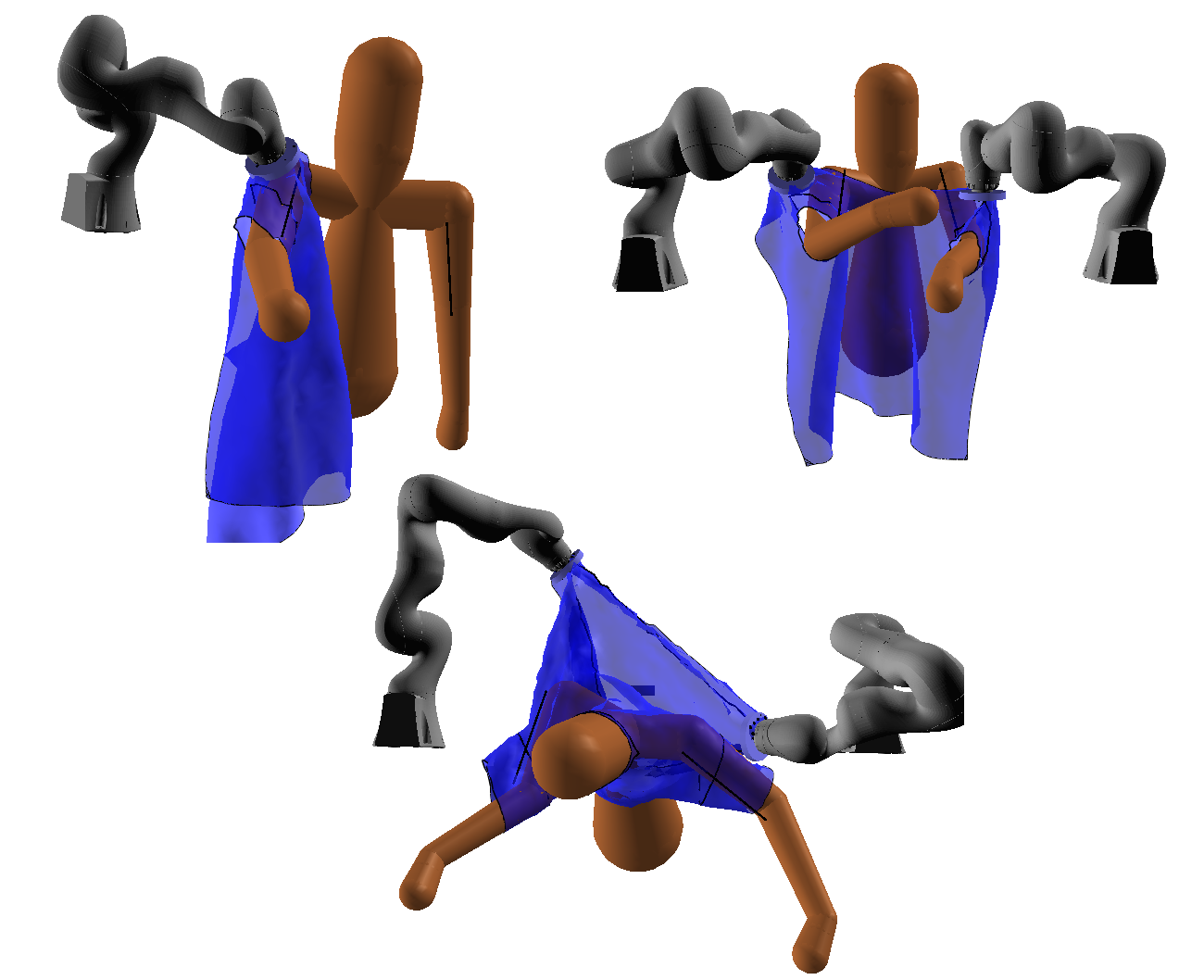}
\caption{Robot-assisted dressing of one and both arms of a hospital gown (top) and a pullover T-shirt (bottom) in simulation.}
\label{fig:assistive_title}
\vspace{-0.5cm}
\end{figure}

Many typical self-dressing tasks can be challenging or insurmountable obstacles for individuals with limited capabilities. For these individuals, family members and/or paid nursing staff often provide daily supplemental care \cite{iom_2008}. To alleviate this burden and enhance independence, personal dressing devices exist on the market today. However, even these require some level of dexterity, strength, and presence of mind to operate and many individuals rely on human assistance. Hence, simulating user capabilities would be valuable for robot learning.


Our work aims to model plausible dressing behavior for both humans and robots in simulated robot-assisted dressing tasks. We make two main contributions in our work. First, we introduce a policy model for human behavior that adapts to variation in a person's physical capabilities.  Second, we demonstrate a co-optimization method to simultaneously train collaborative policies for both a robot and a human using deep reinforcement learning (DRL) and curriculum learning. 


\section{Related Work}

\subsection{Robot-assisted Dressing}

Due in part to nursing shortages and increasing populations of older adults \cite{iom_2008}, robot-assisted dressing has seen a surge of research interest in recent years. Focusing on topological, latent space embedding of cloth/body relationships, Tamei \etal and later Koganti \etal demonstrated robotic dressing of a mannequin's head in a T-Shirt by a two armed robot with the garment beginning on the mannequin's arms \cite{tamei2011reinforcement, 7353860}. Twardon \etal \cite{twardon2018learning} proposed a policy search method to pull a knit hat onto a mannequin's head. Yamazaki \etal applied a trajectory based method for dressing the pants of a passive user with error recovery functions based on vision and force feedback \cite{yamazaki2014bottom}. Work by Gao \etal and Zhang \etal used vision to construct a model of a user's range of motion and a dynamic trajectory plan to dress that user in a vest \cite{7759647, zhang2017personalized}. The I-dress project has proposed several robotic dressing assistance techniques including a learning-from-demonstration approach applied to dressing a jacket sleeve and shoe \cite{pignat2017learning, canal2018joining}.

All of these methods take the approach of directly utilizing existing robotic platforms and sensors. Additionally, they learn or optimize models and control directly on real world data and  expert demonstrations. In contrast, our work applies physics simulations for efficient, risk-free exploration of the task space. Recent work by Kapusta \etal used simulations to search for collaborative dressing plans, consisting of sequences of alternating human and robot actions based on an individuals impairments \cite{kapusta2019TOORAD}. In contrast, we sought collaborative feedback-control policies for robot-assisted dressing that can adapt to a range of impairments and allow the robot and the human to move at the same time.




\subsection{Dressing with Haptics}



Researchers have shown that a time series of end effector velocities and measurements from a gripper-mounted force torque sensor in simulation can be used to both classify sleeve dressing trial success and estimate force maps on the limb being dressed \cite{kapustadata, yu2017haptic, erickson2017how}. The resulting estimators can then be applied to physical systems to enable robotic dressing of a user with garments and static user configurations similar to those seen during training \cite{erickson2018deep, erickson2019multidimensional}. These methods demonstrate the potential for models trained in simulation to enable real robots to assist with dressing in the real world. Our approach applies DRL with simulated capacitive and force sensor observations to discover active collaboration strategies for robots to assist humans with dressing tasks. 

\subsection{Learning to Dress}
We formulate dressing tasks as Markov Decision Processes (MDPs) and solve them using DRL. DRL has been applied largely to rigid-body control tasks in the complex domain of humanoid motor skills \cite{lillicrap2015continuous,schulman2015high, 2018-TOG-DeepMimic}. However, recent work by Clegg \etal has applied Trust Region Policy Optimization (TRPO) \cite{schulman2015trust} to simulated human self-dressing \cite{Clegg2018}. They proposed a set of reward terms and observation features which enabled learning of successful self-dressing control policies for a simulated character dressing garments such as a jacket and T-shirt. Many of these apply to assisted dressing as well. However, DRL for robot-assisted dressing in simulation has not been previously demonstrated with a full robot simulation, or simulated disabilities, nor have such policies been co-optimized for collaboration, or transferred to a real robot. Yet, some prior work has explored concurrent learning (co-optimization) in multi-agent settings.
For example, Gupta et. al. compare concurrent learning to centralized and decentralized approaches for a number of tasks, including a multi-robot problem \cite{gupta2017cooperative}. 

\ignorethis{
\zackory{Cite this paper by Gupta et al.~\cite{gupta2017cooperative}. They use Deep RL for a multi-robot problem with curriculum learning, co-optimization (they call it concurrent learning), and their policies output joint actions.} \zackory{Cite this paper by Chen et al.~\cite{chen2018hardware}. This is a more rigorous study of using joint actions in reinforcement learning (considering hardware constraints) within a multi-robot environment.}}

\section{Methods}

\begin{figure}[t]
\vspace{3mm}
\centering
\includegraphics[width=0.47\textwidth]{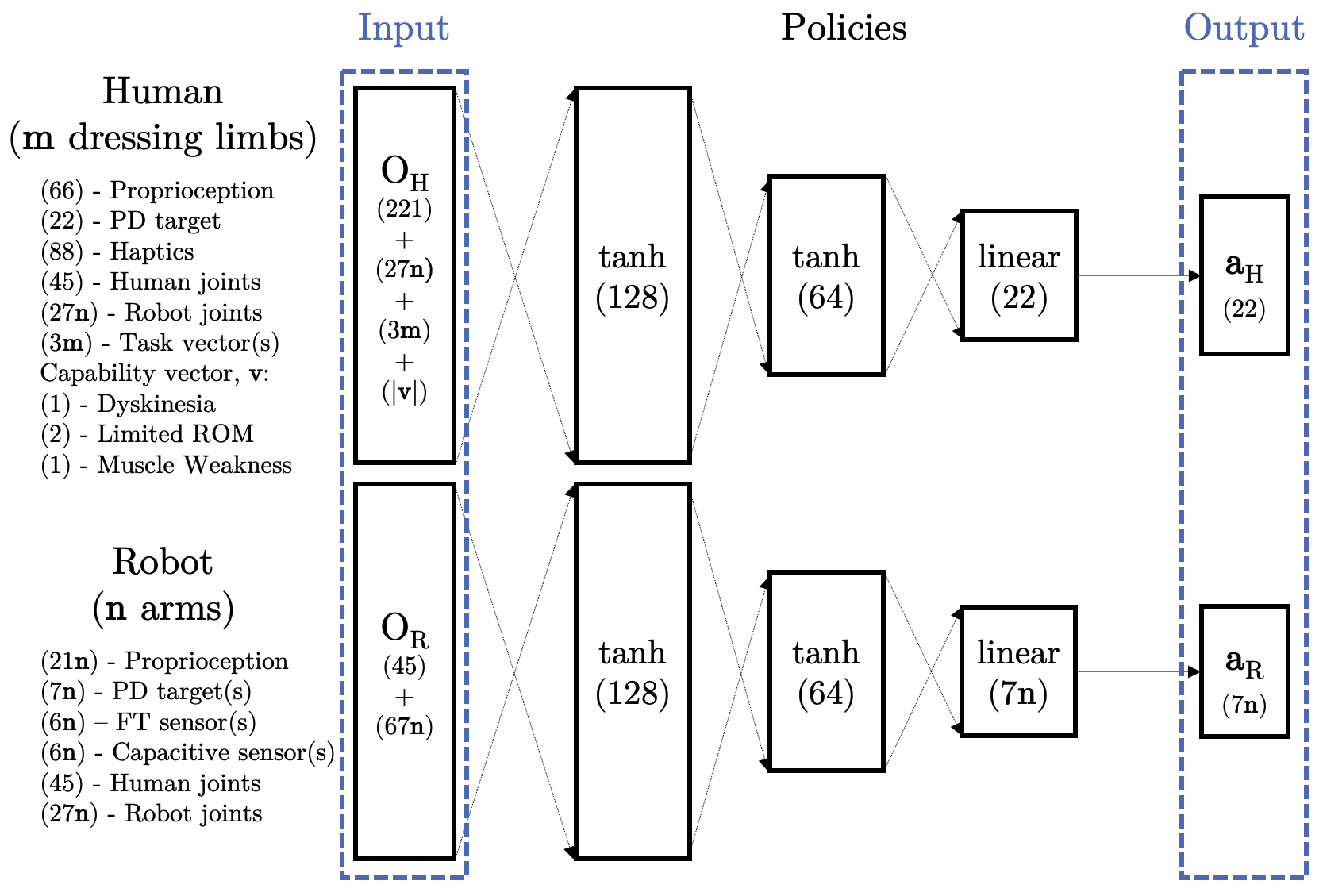}
\caption{Policy network architectures for both robot and human. Input and Output denote the unified observation and action layers  used for co-optimization.}
\label{fig:NN_architecture}
\vspace{-0.5cm}
\end{figure}

We take a co-optimization (concurrent learning) approach to simultaneous training of control policies for the human, $\pi_{\boldsymbol{\theta}_H}$, and robot, $\pi_{\boldsymbol{\theta}_R}$, modeled with fully-connected neural networks as shown in Figure \ref{fig:NN_architecture}. The goal of the optimization is to solve for the policy parameters, $\boldsymbol{\theta}_H$ and $\boldsymbol{\theta}_R$, such that the expected long-term reward is maximized:

\begin{equation}
    \max_{\boldsymbol{\theta}_H, \boldsymbol{\theta}_R} \mathbb{E}[\sum_t r(\boldsymbol{s}_t, \boldsymbol{a}^H_t, \boldsymbol{a}^R_t) ],
\label{eq:cooptimization}
\end{equation}
where $\boldsymbol{s}$ is the full state of the cloth, the human, and the robot, $\boldsymbol{a}^H$ is the action of human and $\boldsymbol{a}^R$ is the action of robot, according to the policies. The reward function, $r$ that measures the task achievement and the cost of actions is jointly defined for the human and the robot. We use a model-free reinforcement learning algorithm, TRPO \cite{schulman2015trust}, to solve for $\boldsymbol{\theta}^H$ and $\boldsymbol{\theta}^R$. We will describe the design of the human policy, the robot policy, and the joint reward function in the remainder of this section.

\subsection{Human Policy: $\pi_{\theta_H}$}
\label{ssec:human_policy}
To model the behavior of the human when receiving dressing assistance, our method solves for a policy that outputs the optimal human action conditioned on the observation made by the human and the motor capability of the human. The action determines the change of the target pose (i.e., target joint angles) which is then tracked by the PD controllers that actuate the human's joints. We use an implicit implementation of PD controllers introduced by Tan \etal \cite{Tan:2011b}. 

The design of the \emph{observation space} of the policy is more challenging. Clegg \etal \cite{Clegg2018} proposed an observation space for learning a human self-dressing policy, including information about proprioception, haptics, surface information, and the task. We extend their observation space for robot-assisted dressing tasks by adding the 3D position of all robot and human joint positions, $\mathcal{O}_{jp}$, and the target pose from the previous time step, $\mathcal{O}_{tar}$. 

In addition to the human's observation, $\pi_{\theta_H}$ is also conditioned on the motor capability of the human. This additional input to the policy enables a single neural network to model a variety of human capabilities within the domain of dressing tasks. We define a capability vector $\boldsymbol{v}$, each element of which encodes one dimension of human capability, such as muscle strength or joint limits, as input to the policy $\pi_{\theta_H}$. With this formulation, each additional dimension of variation in capability can be incorporated by appending an element to $\boldsymbol{v}$ to identify it. We demonstrate this approach by simulating three distinct capability variations: dyskinesia, limited range of motion, and muscle weakness.

\subsubsection{Dyskinesia}
Dyskinesia is a category of motion disorders characterized by involuntary muscle movements that has been linked to functional difficulties with activities of daily living \cite{pahwa2018impact}. Our simple model of dyskinesia adds random noise drawn uniformly from a pre-defined range to the PD target pose. The magnitude of the noise is then included in the capability vector $\boldsymbol{v}$. Note that this noise does not affect the target pose $\mathcal{O}_{tar}$ as input to the policy. Instead, it directly modifies the target pose of the PD controllers, resulting in torques that do not precisely track the desired motion.

\subsubsection{Limited Range of Motion}
Common injuries, diseases, and disorders can result in temporary or permanent limitations to joint mobility. The resulting motion constraints can add additional challenge to dressing tasks and necessitate adaptive strategies. We modeled this by modifying both the upper and lower limits of a joint (e.g. the elbow). By doing so, we introduced two dimensions of variation, $j_{min}$ and $j_{max}$ as a part of capability vector $\boldsymbol{v}$. Both $j_{min}$ and $j_{max}$ can be sampled from separate ranges with $j_{max} > j_{min}$ to enforce physicality.

\subsubsection{Muscle Weakness}
We modeled muscle weakness by applying a scaling factor to the torque limits of the human. The scaling factor is sampled uniformly from a pre-defined range and is included as a type of capability variation in the capability vector $\boldsymbol{v}$.

\subsection{Robot Policy: $\pi_{\theta_R}$}
\label{ssec:robot_policy}
Our approach aims to discover assistance strategies which could be applied to a physical robotic system and would accommodate transfer of the policy to a real robot. As such, we considered that individual features in the observation space should be reasonably obtained by a physical system. Similar to the human policy, the action space is defined as the change of the target pose (i.e., target joint angles) tracked by the robot's PD controllers. The observation space includes the robot's proprioception, readings from end-effector mounted force-torque and capacitive sensors, the 3D positions of robot and human joints ($\mathcal{O}_{jp}$), and the target pose from the previous time step. All robot arms share both observation and action space within one policy and can therefore be considered together as a single robotic system. Each term in the observation space with the exception of human joint positions is provided to the policy for each robot arm in the system. This allows a simple mechanism for expanding the system to include additional manipulators.

The robot's proprioception (i.e., joint angles and velocities) is  provided by the current simulated state of the robot. However, we augmented the physics simulator to model force-torque and capacitive sensors.

\subsubsection{Force-torque Sensor}\label{ssec:ft_sensor}
\begin{figure}[t]
\vspace{3mm}
\centering
\includegraphics[width=0.4\textwidth]{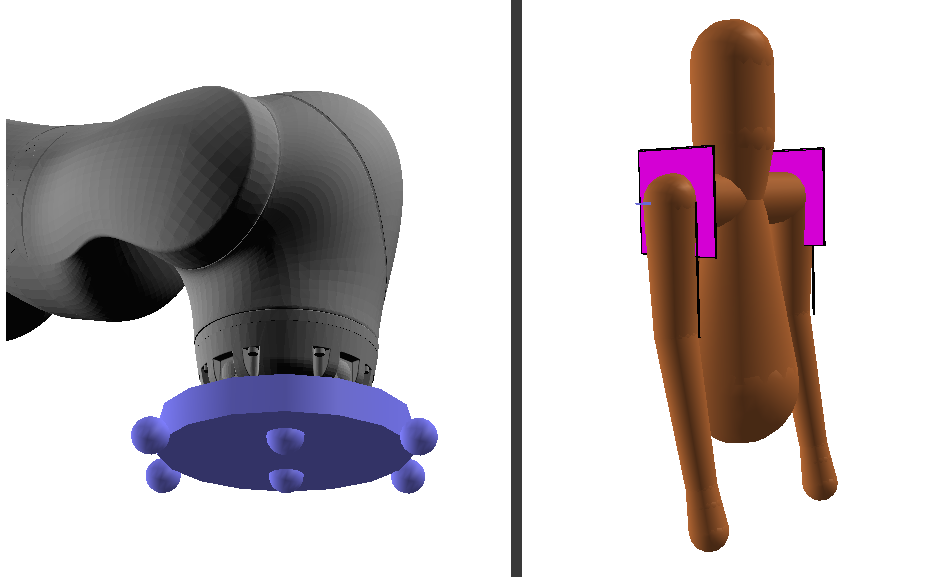}
\caption{The end-effector and cylindrical gripping tool of a simulated KUKA LBR iiwa arm with a 2x3 grid of capacitive sensor locations displayed (left). The default rest pose and sleeve dressing success thresholds (purple planes) for our simulated human model (right).}
\label{fig:cap_and_threshold}
\vspace{-0.5cm}
\end{figure}

An end-effector mounted force-torque sensor is a common choice for robot control tasks. Such a sensor can be attached to the robot after its most distal joint and before any specialized tool such as a cloth or object gripper. In the domain of robot-assisted dressing alone, force-torque sensors have been used for error detection, simulated and real state estimation, and model predictive control \cite{yamazaki2014bottom, zhang2017personalized, erickson2017how, erickson2018deep, kapustadata, yu2017haptic}. This sensor enables a robot holding a garment to monitor the force it applies to the human by pulling the garment or by directly making contact with its end effector. We compute garment forces from local deformation of the cloth in the neighborhood of mesh vertices gripped by the robot. Rigid contact forces between the cylindrical gripping tool shown in Figure \ref{fig:cap_and_threshold} (left) and the human are also detected. The resulting observation feature for each robot is the six dimensional sum of these force and torque vectors.

\subsubsection{Capacitive Sensor}
In many human robot interaction tasks, vision systems are used to estimate the state of the human in order to plan or adapt control strategies. However, clothes often occlude the body, making the assisted dressing task challenging for a purely vision based approach. Capacitive sensors, however, are not limited by visibility and recent research in robot-assisted dressing has demonstrated their use for detecting proximity to the human body through clothing \cite{8307487}. We simulated a 2x3 grid of capacitive sensors on the garment gripping surface of the robot end-effector. Following experiments by Erickson \etal we modeled each sensor in simulation as a proximity detection sphere with a 15cm range \cite{erickson2019multidimensional}. We computed the scalar distance reading for each sensor by finding the closest point to the sensor on the human body model and returning the distance to that point clipped to 15cm. The resulting observation feature for each robot is the six dimensional concatenation of these sensor readings. Figure \ref{fig:cap_and_threshold} (left) shows the placement of these sensors on the simulated robot end-effector.

\subsection{Joint Reward Function}
\label{ssec:joint_reward}
Our proposed co-optimization approach allows simultaneous training of the robot and human control policies. This formulation necessitates a unified reward function for the dressing task. While the input to $\pi_{\theta_H}$ and $\pi_{\theta_R}$ should be restricted by the capability of human and robot perception, the input to the reward function can take advantage of the full state of the simulated world $\boldsymbol{s}$, because the reward function is only needed during training time.

Previously, Clegg \etal proposed a set of reward terms for the self-dressing task, including a progress reward, deformation penalty, geodesic reward, and per-joint rest pose reward: $[r_{p}(\boldsymbol{s}), r_{d}(\boldsymbol{s}), r_{g}(\boldsymbol{s}), \vc{r}_{r}(\boldsymbol{s})]$ \cite{Clegg2018}. Directly using this reward function for robot-assisted dressing tasks resulted in robots learning aggressive strategies that sacrifice safety for task progress. For example, we observed the robot stretching the garment against the human's neck or using large forces to shove the human. 

As such, we extended the above reward function with an additional term and some modifications. We introduced a penalty on force perceived by the human from both the garment and the robot, $r_{c}(\boldsymbol{s})$. The perceived force penalty is a function of the magnitude of the largest aggregated force vector, $f_{max}(\boldsymbol{s})$, applied on the human in the current state $\boldsymbol{s}$:
\begin{equation}
    r_{c}(\boldsymbol{s}) = -\frac{\tanh(w_{scale}(f_{max}(\boldsymbol{s}) - w_{mid}))}{2} - \frac{1}{2},
\end{equation}
where $w_{mid}$ defines the midpoint of the penalty range and $w_{scale}$ scales the slope and upper/lower limits of the penalty function. These parameters were chosen empirically based on the range of typical contact forces during dressing tasks. 

We modified the rest pose reward, $\vc{r}_{r}(\boldsymbol{s})$, which penalizes human poses deviating from its rest pose to include a per-joint weight vector, $\vc{w}_5$. This reward term serves as an important regularizer when the human is being dressed. The values of $\vc{w}_5$ must be set carefully, as any particular setting will result in the human policy prioritizing use of some joints over others. We also modified the progress reward, $r_{p}(\boldsymbol{s})$, to include a success threshold as shown in Figure \ref{fig:cap_and_threshold} (right). Limb progress beyond this threshold results in maximum reward and a bonus reward equivalent to an additional 100\% of the progress reward weight. This modification prioritizes task success while reducing the occurrence of unsafe progress reward maximization strategies such as stretching the sleeve against the neck or underarm.

The complete reward at state $\boldsymbol{s}$ is computed as $r(\boldsymbol{s})= w_1 \cdot r_{p}(\boldsymbol{s}) +w_2 \cdot r_{d}(\boldsymbol{s})+ w_3 \cdot r_{g}(\boldsymbol{s}) + w_4 \cdot r_{c}(\boldsymbol{s}) + \vc{w}_5 \cdot \vc{r}_{r}(\boldsymbol{s})$ where the scalar weights, $w_{1-4}$, and the rest pose weight vector, $\vc{w}_5$, are determined empirically for each task.

\subsection{Refining Policies with Curriculum Learning}
Our joint reward function, $r(\boldsymbol{s})$, contains several competing objectives. The human and robot are strongly rewarded for making progress on the dressing task and penalized for deforming the garment and colliding with one another. The optimization process described by equation \ref{eq:cooptimization} is sensitive to the balance of reward weights. Strong penalty weights often stifle exploration early in the process and can prevent the discovery of successful strategies altogether. Reducing these penalty weights, however, will often result in policies which succeed at the primary task, but do not satisfy secondary requirements.

Curriculum learning strategies provide one possible solution to this problem and have been successfully applied to deep reinforcement learning of motor skills tasks in the past \cite{Bengio:2009,karpathy2012curriculum}. In our case, assisted dressing policies trained with strongly weighted penalties on the magnitude of force perceived by the human (\ie large $w_4$) failed to succeed at the task. This approach resulted in extremely sub-optimal policies in which the robot and human avoided exploring the task due to large penalties for accidental contact. However, decreasing $w_4$ resulted in successful policies which caused the human and robot to interact in unsafe ways. We applied a curriculum learning approach to reduce unwanted contact forces by first training successful dressing policies with a low value of $w_4$ and then refining these policies with an increased $w_4$ penalty.

\section{Implementation}\label{sec:implementation}

To demonstrate our approach for simulating robot-assisted dressing, we chose to examine three scenarios: dressing a single arm of a hospital gown, dressing two arms of a hospital gown, and dressing a pullover T-shirt. The following sections describe the implementation details for each of these examples with reward weights set as in table \ref{tbl:reward_weights}. In addition, we strongly recommend that the reader view the accompanying video for visualization of these results.

We simulated the physical interactions between the human and robot with the Dynamic Animation and Robotics Toolkit (DART) \cite{DART} at 400Hz. We included an additional data driven joint limiting constraint demonstrated by \cite{jiang2017data} for the shoulders and elbows of the human to more accurately model the range of motion for these joints. Additionally, we used the inverse kinematics feature of PyBullet \cite{coumans2019} to compute the initial poses for each robot arm from sampled end effector transforms. We simulated cloth dynamics at 200Hz with NVIDIA PhysX \cite{physx}. A pre-simulation phase of half a second is used to move the garment into sampled grip positions at the start of each episode.

All policies are represented by fully connected neural networks consisting of input followed by 2 hidden layers of 128 and 64 nodes respectively with tanh activations and a final linear layer before output as shown in Figure \ref{fig:NN_architecture}. During training the input and output layers for both robot and human policies are concatenated to enable co-optimization. However, in all cases the robot and human control policies are separate and share no connections. Policies were trained for between 700 and 1500 TRPO iterations with $40,000$ policy queries per iteration and $600$ queries per episode. Articulated rigid body physics simulation using DART were run at $400$Hz and cloth simulation using PhysX at $200$Hz with the policies queried at $100$Hz such that each episode consists of 6 seconds of simulated time. With the addition of PD control and sensor simulation, our environment produces a sample in 0.05-0.15 seconds on a 4GHz AMD FX-8350 (without multi-threading). Co-optimization of robot and human policies for a single task requires about 24 hours of simulation time on the AWS EC2 c5.9xlarge (36 vCPUs, 3 GHz, Intel Xeon Platinum 8124M) compute nodes used.

\begin{figure}[t!]
\vspace{3mm}
\centering
\includegraphics[width=0.47\textwidth]{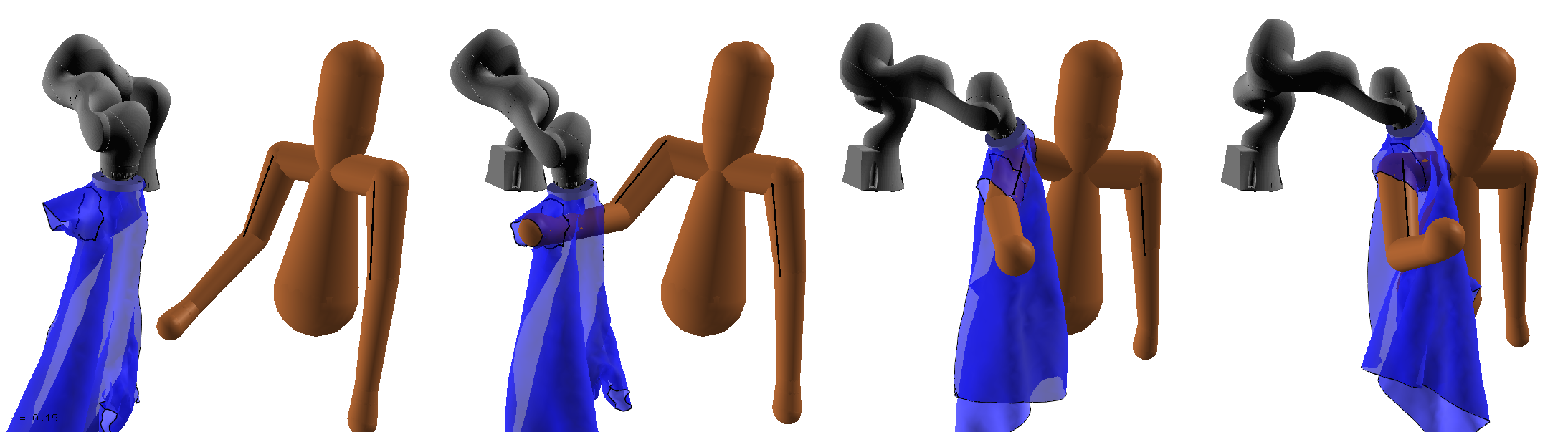}
\caption{A robot assists a fully capable human in dressing one sleeve of a hospital gown in simulation.}
\label{fig:typical_onearm_gown_sequence}
\vspace{-0.5cm}
\end{figure}

\subsection{Dressing One Arm in a Hospital Gown}
In this task, the human and robot work together to dress the human's right arm in the sleeve of the hospital gown (Figure \ref{fig:typical_onearm_gown_sequence}). The robot grips the sleeve of a hospital gown at an initial position drawn from a uniform random distribution within reach of the robot, in front of the human, and on the robot's right. We applied the capability models described in \ref{ssec:human_policy} to this task. The capability vector of the human policy is randomly sampled at the initialization of each episode, but the robot policy is given no indication of this capability and must learn to infer it from its other observations.

\subsubsection{Unilateral Dyskinesia}
We applied our model of dyskinesia to the active (i.e. dressing) arm. The magnitude of noise applied to each DOF in the affected joints is sampled uniformly with a maximum deviation of $n_{max}$ percent of that DOF's range of motion from the target joint angle. The maximum deviation, $n_{max}$, is drawn uniformly from the range [0, 15] percent upon initialization of each episode and the normalized observation $\frac{n_{max}}{15}$ is included in the capability vector of the human policy's input observation.

\subsubsection{Elbow Joint Limit Constraint}
We modeled the variation of elbow joint limits by uniformly sampling the upper and lower bounds of the elbow range upon initialization of each episode. We allowed for a wide range of variation from no elbow limitation to severe limitation where the range is restricted to a single pose.

\subsubsection{Unilateral Muscle Weakness}
We modified the task to introduce unilateral weakness of the active (i.e. dressing) arm. We trained the human and robot policies on the torque scaling range [0.1, 0.6]. The lower end of this capability variation range is not completely incapable (i.e. scaling by 0) as we do not expect the robot to physically manipulate the human during assistance and therefore the human must be capable of some active participation. However, a human with only 60\% strength is still more than capable of raising its arm and inserting it into a sleeve. Samples in the range [0.6, 1.0] resulted in a heavy bias towards capable humans and fewer opportunities for the policies to learn in the presence of impairment. The upper end of the range [0.1, 0.6] was chosen to limit this bias toward more capable humans.

\begin{figure}[t!]
\vspace{3mm}
\centering
\includegraphics[width=0.49\textwidth]{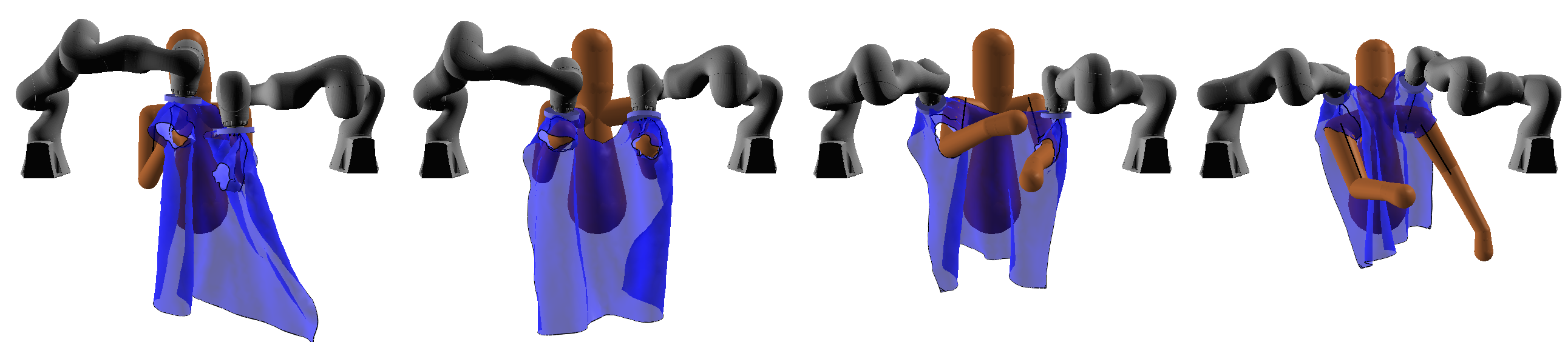}
\caption{A robot assists a fully capable human in dressing both sleeves of a hospital gown in simulation.}
\label{fig:typical_twoarm_gown_sequence}
\vspace{-0.5cm}
\end{figure}

\subsection{Dressing Both Arms in a Hospital Gown}
To demonstrate the ability of our proposed approach to generalize to tasks which involve more than one robot, we designed a two-arm hospital gown dressing task. In this task, two robots are positioned in front and on each side of the human. Each robot grips a sleeve of hospital gown at initial positions drawn from a uniform random distribution within reach of the robots, in front of the human, and with distance greater than 0.1 meters and less than 0.5 meters between robot end effectors.

 One dressing progress reward term per sleeve is provided for this task. The human observation space is expanded to include additional observations for the second arm. The human is initialized with elbows bent to start this task. Figure \ref{fig:typical_twoarm_gown_sequence} shows a sequence of frames demonstrating the results of applying our technique to the two arm hospital gown task.

\subsection{Dressing a Pullover T-shirt}
Our approach can also be applied to more complex dressing tasks. We demonstrate this by training human and robot control policies for a pullover T-shirt dressing task. Starting with bent elbows, the human must insert both arms and its head into the shirt being pulled on by two robot arms. The robots are in the same position as the two arm gown example and have the same end effector initialization volumes.

In this task, the human has three active dressing limbs: two arms and the head. All three provide individual observations to the policy and reward terms to $r(\textbf{s})$. The arms are rewarded for progress with respect to the sleeves, while the head is rewarded for progress with respect to the collar. All three are additionally rewarded for progress with respect to the waist of the T-shirt to encourage improvements early in training. This is a challenging task for a single pair of policies to accomplish and took two to three times as long to train. 

We applied an exit criterion to demonstrate these controllers. Once all three limbs are detected as dressed, control transitions from the trained neural networks to a feed-forward pose tracking controller. At this point, the robot's grip constraints on the garment are released. Readers are encouraged to view this result in the accompanying video.

This task relates to past work at the Nara Institute of Science and Technology \cite{tamei2011reinforcement, 7353860} demonstrating a robot pulling a T-shirt over the head of a mannequin with arms already in the sleeves. Our approach discovered a similar strategy to that which was previously defined and implemented manually.


\begin{table}[t!]
\vspace{0.25cm}
\caption{Reward weights for various task implementations \newline $w5$: (torso, spine, neck, left arm, right arm) }
\vspace{-0.25cm}
\label{tbl:reward_weights}
\begin{center}
\begin{tabularx}{0.483\textwidth}{|X|c|c|c|c|c|c|c|c|c|}
\hline
Task & $w1$ & $w2$ & $w3$ & $w4$ & \multicolumn{5}{c|}{$w5$}\\
\hline
\hline
Gown: & 40 & 5 & 0 & 5 & 40 & 4 & 8 & 4 & 0.5 \\
One Arm & & & & & & & & & \\
\hline
Gown: & 40 & 5 & 0 & 5 & 45 & 5 & 4 & 0.25 & 0.25\\
Two Arms & & & & & & & & & \\
\hline
T-shirt & 20 & 5 & 15 & 5 & 40 & 5 & 4 & 0 & 0\\

\hline
\end{tabularx}
\end{center}
\vspace{-0.5cm}
\end{table}

\section{Evaluation} \label{sec:assisted_evaluation}

To evaluate our approach, we conducted a series of experiments tracking dressing progress and maximum perceived contact force on the human's body. We applied our trained control policies to 100 episodes of several dressing tasks with parameters sampled randomly from the training distributions. We evaluated success rates on the one arm hospital gown task with each of the described capability variation types and on the two arm hospital gown task with a fully capable human. We define success as limb progress passing the defined threshold for success described in section \ref{ssec:joint_reward} before the end of the six second episode horizon.

In order to maximize reward, trained policies often exhibit high speed dressing strategies which may be undesirable if executed on physical robot hardware. To address this, we examined the results of scaling policy outputs in order to manipulate task execution speed without re-training. We then compared contact forces before and after applying our curriculum learning approach. Finally, we conducted an ablation study on the robot's observation of capacitive sensor readings and human joint positions.

\subsection{Success Rates Across Tasks}
\subsubsection{High Success Rates Achieved without Impairments}
We first applied our approach to the one arm hospital gown dressing task with a fully capable human, resulting in a 100\% success rate. The two arm hospital gown task appears more challenging with a 100\% success rate of the right arm, but a 96\% success rate of the left arm. During training, the right arm was the first to be successful and the policies tend to dress that arm fully despite miss cases of the left arm. 

\subsubsection{Training with Impairments Improved Performance}
\begin{figure}[b!]
\vspace{-0.4cm}
\centering
\includegraphics[width=0.45\textwidth]{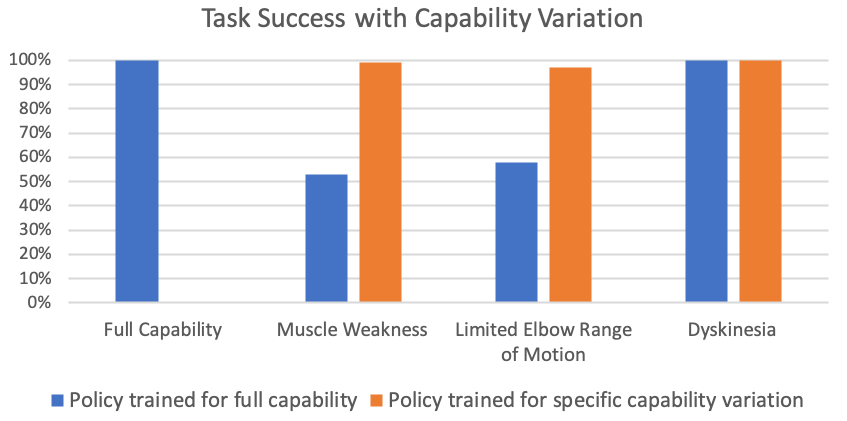}
\vspace{-0.25cm}
\caption{Comparison of success rates on one arm hospital gown dressing task with capability variations for policies trained on a fully capable human (left) and policies trained on specific variation ranges (right).}
\label{fig:capability_variations}
\end{figure}

We then evaluated the relative difficulty of individual capability variations and the generality of policies trained without variable capability. To do so, we tested control policies trained for the fully capable human on the one arm dressing task for each capability variation. We found that the control policies for the capable human are 100\% successful with dyskinesia, demonstrating that controllers trained with our approach are able to generalize surprisingly well to noisy human actions. The 53\% success rate on muscle weakness variations, and 57\% success rate with limited elbow mobility demonstrate the challenge of adapting to unseen impairment of certain capabilities. We then compared the performance of these policies to that of policies trained specifically for each capability range with our approach. As shown in Figure \ref{fig:capability_variations}, training the policies with impairment using our approach allowed them to discover successful collaboration strategies for robot-assisted dressing with variable human capability.

\subsubsection{Success Rates Depend on Impairment Range}
\begin{figure}[b!]
\vspace{-0.2cm}
\centering
\includegraphics[width=0.45\textwidth]{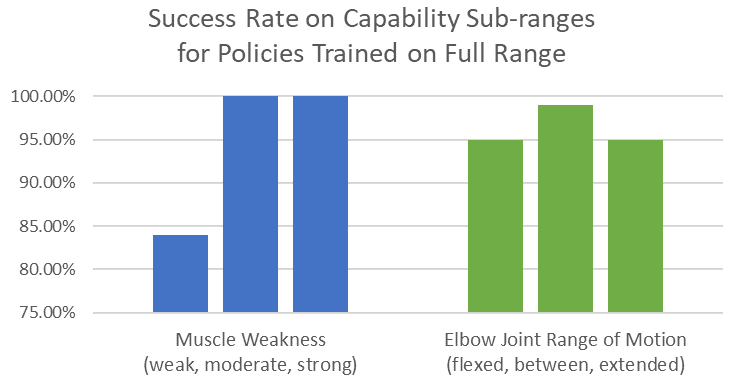}
\caption{Comparison of success rates for one arm hospital gown dressing task on subdivided capability ranges for policies trained on the full capability range.}
\label{fig:capability_subdivision}
\end{figure}

We further examined the results of this experiment by dividing both the muscle weakness and elbow joint limit variation ranges into three equally sized sub-ranges and applying our specifically trained control policies to 100 episodes of each sub-range. The results of this experiment are shown in Figure \ref{fig:capability_subdivision} and clearly show that the weakest sub-range of the muscle weakness impairment is the most challenging. The low rate of success is likely the result of sample bias favoring more capable humans during training. In the case of varying elbow range of motion, more capable humans fall into the middle range and we see a normal curve: 95\%, 99\%, 95\%. The most challenging cases here are those with very low joint ranges (high stiffness) and a fully bent or extended arm. With the arm fully extended, the robot must catch the human's hand and wrist much farther from the body. In contrast, a fully bent arm necessitates a tight turn close to the body without catching the cloth on the elbow. These experiments show that care should be taken with capability distributions which contain relatively small regions of high difficulty.

\subsection{Scaling Actions Results in Slower Success}\label{ssec:action_scaling}
Most physical assistive robots at this time move slowly and with a high degree of compliance for their own safety and that of nearby humans and objects. As our results have shown, control policies acquired through our approach move much faster than we may desire outside of simulation. This is in part due to the reward function which prioritizes dressing progress over anything else, pushing both the robot and the human to complete the task as quickly as possible. Since our control policies output the desired change in target pose, one method for reducing control velocity is to directly scale the control outputs, resulting in a reduced rate of change for target positions. Table \ref{tbl:action_scale} compares the success rate and average time to success of several choices of action scale over 100 fixed random seeds. We observed that success rate decreased slightly while average time to success increased significantly. This may indicate that while quick reactive control is necessary in some instances, an action scaling approach to task velocity reduction may often suffice.

\begin{table}[t!]
\vspace{0.25cm}
\caption{Results of scaling actions produced by a trained policy}
\vspace{-0.4cm}
\label{tbl:action_scale}
\begin{center}
\begin{tabular}{|c|c|c|}
\hline
Action Scale & Success Rate & Avg. Time to Success\\
\hline
\hline
1.0x & 100\% & 1.53s\\
\hline
0.8x & 98\% & 1.43s\\
\hline
0.6x & 98\% & 1.65s\\
\hline
0.4x & 97\% & 2.20s\\
\hline
0.2x & 94\% & 3.77s\\
\hline
\end{tabular}
\end{center}
\end{table}

\subsection{Curriculum Learning Lowers Applied Forces}\label{ssec:curriculum_evaluation}
We evaluated the effectiveness of our curriculum training approach by comparing policies trained before and after curriculum learning is applied. During curriculum training, we introduced a linear penalty on the magnitude of maximum contact force perceived by the human at each state in an episode of the task. We first attempted training a new policy from random initialization with the addition of this penalty term. The resulting policy achieved 0\% success, instead finding a local minimum with the robot keeping the garment just out of reach to avoid potential collision states. We then applied our curriculum learning strategy by fine tuning existing robot and human control policies with the same contact force penalty. Table \ref{table:curriculum} shows the resulting reduction in episode maximum perceived contact forces for various tasks. While this approach effectively reduces high contact forces in most cases, it remains clear that further research on contact force reduction should be undertaken.

\subsection{Multimodal Sensing Improves Performance}\label{ssec:robot_ablation_evaluation}
In section \ref{ssec:robot_policy}, we introduced a set of observation features for the assistive robot policy. The robot acquires information about the human state through: force-torque sensor readings, capacitive sensor readings, and the joint positions of the human. We consider inclusion of the force-torque sensor to be important for the safety of the human and therefore do not consider ablating it. However, observation of the capacitive sensor readings and human joint positions serve similar functions and are candidates for removal.

\begin{table}[t!]
\vspace{-0.25cm}
\caption{Percentage of episodes with maximum contact forces below 50 Newtons before and after curriculum refinement}
\vspace{-0.3cm}
\label{table:curriculum}
\begin{center}
\begin{tabular}{|c|c|c|}
\hline
Trial & Before Curriculum & After Curriculum\\
\hline
\hline
Hospital Gown: One Arm & - & -\\
\hline
Full Capability & 9\% & 95\% \\
\hline
Limited Range of Motion & 19\% & 95\% \\
\hline
Muscle Weakness & 33\% & 65\% \\
\hline
\hline
Hospital Gown: Two Arms & 2\% & 91\% \\
\hline
\end{tabular}
\end{center}
\vspace{-0.7cm}
\end{table}

 We conducted an ablation study on these observations for the unilateral muscle weakness variation without curriculum learning. Without the capacitive sensor, we observed a drop in success rate from 99\% to 79\%. In this case, the robot must learn the complex relationship between geometry and joint positions in order to avoid contact with the human and pull the sleeve onto the arm. Without the human joint position feature, we observed a reduced 75\% success rate. Likely, this is the result of the robot observing the human only through its end effector mounted capacitive sensor and being, therefore, unaware of the true state of the limb. Both ablations resulted in significant increases in contact forces perceived by the human from both the garment and rigid contact. These results suggest that both observations provide important localization information during difficult interaction tasks.
 
\subsection{Policies Trained in Simulation can Transfer to Real World}\label{ssec:sim_to_real}
We conducted a preliminary evaluation of learned policies on a real robot. For safety, we modeled a single arm dressing task where a real PR2 robot dressed the sleeve of a hospital gown onto the arm of a humanoid Meka robot. We trained only the assistive PR2 policy with the Meka robot attempting to hold a fixed pose while affected by our involuntary motion model (dyskinesia).

\begin{figure}[t!]
\vspace{1mm}
\centering
\includegraphics[width=0.47\textwidth, height=3cm]{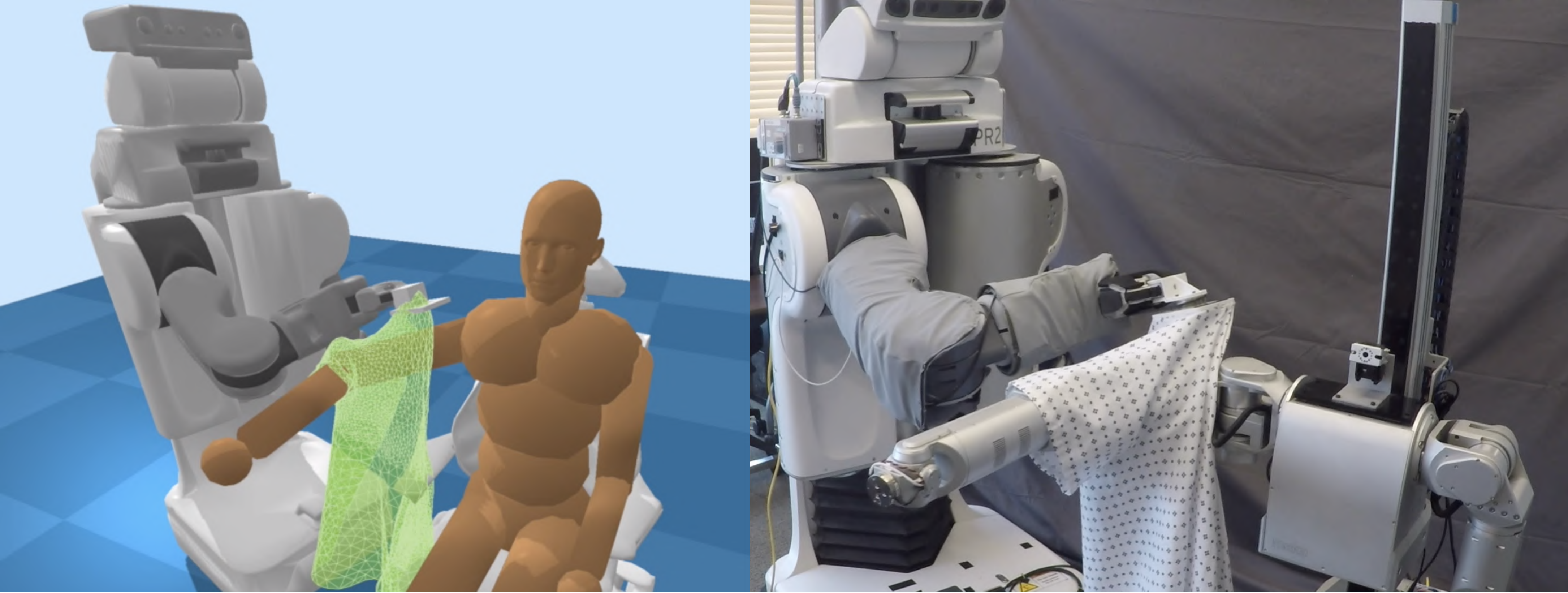}
\vspace{-0.25cm}
\caption{The PR2 dresses the sleeve of a Meka robot emulating a human with dyskinesia holding a fixed arm pose with a 45 degree bend in the elbow in simulation (left) and reality (right).}
\label{fig:sim_to_real}
\vspace{-0.5cm}
\end{figure}

To demonstrate the ability of our proposed approach to generalize to other simulation frameworks, we used the Assistive Gym framework for training in simulation (using Proximal Policy Optimization) \cite{erickson2019assistive}. Similar to \cite{yu2017haptic} we used CMA-ES to optimize the simulated cloth parameters in Assistive Gym to better match the applied forces and torques during a manually controlled dressing trial between the PR2 and the Meka humanoid. These cloth parameters included the cloth stiffness, damping, and friction. This calibration of the simulator was our only use of real world data. 

We also modified parameters from Section \ref{sec:implementation}, decreasing the action rate to 10Hz and increasing episode length to twenty seconds to accommodate the capabilities of the physical system. As described in section \ref{ssec:action_scaling}, we slowed the trained policy's execution on the robot by scaling its actions by $0.4$x. These changes resulted in the additional benefit of slower and safer policies both in simulation and in reality. Due to the use of a robotic stand-in for the human, we chose to omit the capacitive sensor in place of additional reward terms penalizing distance and orientation discrepancies between the arm and the robot's end effector. Other reward terms and observation features remained unchanged.

Through this experiment we were able to demonstrate successful transfer of neural network policies trained from random initialization via DRL in simulation to a physical robotic system as shown in Figure \ref{fig:sim_to_real}. We encourage the reader to view our video results for a visual comparison of this transfer. Over 20 trials dressing a straight arm in reality, policies trained in simulation achieved 90\% success with the magnitude of the forces measured by a 6-axis force-torque sensor at the robot's end effector being in the range [0~N, 8.35~N] with an average of 1.02~N.
In addition, we trained a policy to dress a hospital gown sleeve onto a person with a 45$^{\circ}$ bend in their elbow. When evaluated on the physical PR2 and Meka robots, as seen in Figure \ref{fig:sim_to_real}, this policy achieved 100\% success over 20 dressing trials, with all measured forces in the range of [0~N, 3.33~N] and an average of 0.66~N.
The success of policies trained in a second simulation framework and their subsequent transfer to reality serves as evidence for the generality of our approach.

\section{Conclusion}
We have presented a DRL based approach for modeling collaborative strategies for robot-assisted dressing tasks in simulation. Our approach applies co-optimization to enable distinct robot and human policies to explore the space of joint solutions in order to maximize a shared reward. In addition, we presented a strategy for modeling impairments in human capability. We demonstrated that our approach enables a robot, unaware of the exact capability of the human, to assist with dressing tasks in simulation. These policies are credible examples of human-robot collaboration that can provide insights into robot-assisted dressing. Our results indicate that action scaling can reduce execution speed, curriculum learning can be used to improve safety, and multimodal sensing improves performance. Finally, we conducted a simulation to reality transfer experiment demonstrating one robot successfully dressing another by applying a policy trained with our approach in simulation.

\bibliographystyle{IEEEtran}
\bibliography{references}

\end{document}